\documentclass{article}

\usepackage{arxiv}

\usepackage[utf8]{inputenc} 
\usepackage[T1]{fontenc}    
\usepackage{hyperref}       
\usepackage{url}            
\usepackage{booktabs}       
\usepackage{amsmath, amsfonts}       
\usepackage{nicefrac}       
\usepackage{microtype}      
\usepackage{cleveref}       
\usepackage{lipsum}         
\usepackage{graphicx}
\usepackage{natbib}
\usepackage{doi}

\title{A Hybrid Architecture for Out of Domain Intent Detection and Intent Discovery}

\date{}

\author{ \href{https://orcid.org/0000-0002-7691-0778}{\includegraphics[scale=0.06]{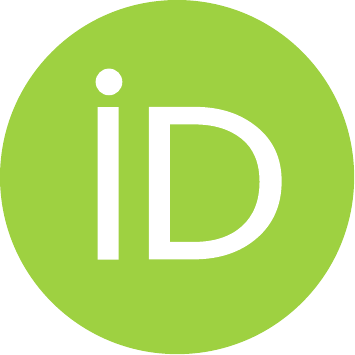}\hspace{1mm}Masoud Akbari} \\
	Department of Computer Science\\
	Amirkabir University of Technology\\
	\texttt{ma.akbari421@aut.ac.ir} \\
	\And
	{Ali Mohades} \\
	Department of Computer Science\\
	Amirkabir University of Technology\\
	\texttt{mohades@aut.ac.ir} \\
	\And
	{M. Hassan Shirali-Shahreza} \\
	Department of Computer Science\\
	Amirkabir University of Technology\\
	\texttt{hshirali@aut.ac.ir} \\
}

\renewcommand{\headeright}{}
\renewcommand{\undertitle}{}
\renewcommand{\shorttitle}{A Hybrid Architecture for Out of Domain Intent Detection and Intent Discovery}

\hypersetup{
pdftitle={A template for the arxiv style},
pdfsubject={q-bio.NC, q-bio.QM},
pdfauthor={David S.~Hippocampus, Elias D.~Striatum},
pdfkeywords={First keyword, Second keyword, More},
}

\begin{document}
\maketitle

\begin{abstract}
	Intent Detection is one of the tasks of the \textbf{N}atural \textbf{L}anguage \textbf{U}nderstanding (NLU) unit in task-oriented dialogue systems.
  \textbf{O}ut \textbf{o}f \textbf{S}cope (OOS) and \textbf{O}ut of \textbf{D}omain (OOD) inputs may run these systems into a problem.
  On the other side, a labeled dataset is needed to train a model for Intent Detection in task-oriented dialogue systems. 
  The creation of a labeled dataset is time-consuming and needs human resources. 
  The purpose of this article is to address mentioned problems. 
  The task of identifying OOD/OOS inputs is named OOD/OOS Intent Detection. 
  Also, discovering new intents and pseudo-labeling of OOD inputs is well known by Intent Discovery. 
  In OOD intent detection part, we make use of a Variational Autoencoder to distinguish between known and unknown intents independent of input data distribution. 
  After that, an unsupervised clustering method is used to discover different unknown intents underlying OOD/OOS inputs. 
  We also apply a non-linear dimensionality reduction on OOD/OOS representations to make distances between representations more meaning full for clustering. 
  Our results show that the proposed model for both OOD/OOS Intent Detection and Intent Discovery achieves great results and passes baselines in English and Persian languages.
\end{abstract}

\keywords{Natural Language Understanding \and Out-of-Domain Intent Detection \and Intent Discovery}

\section{Introduction}
Natural Language Processing is a set of computational methods
that tries to process human language in different applications using
linguistic analysis \citep{liddy2001natural}. With the advancement of deep learning
approaches in recent years, research on NLP is developing rapidly.
Studies related to NLP are divided into many categories: question answering, text summarization,
topic modeling, sentiment analysis, etc \citep{eisenstein2019introduction}.
Among all these usages, task-oriented chatbots are a part of these categories that have taken much attention.
Generally, these kinds of chatbots consist of 3 central units:
\textbf{N}atural \textbf{L}anguage \textbf{U}nderstanding (NLU), Dialogue Management, and \textbf{N}atural \textbf{L}anguage \textbf{G}eneration (NLG) \citep{galitsky2019chatbot}. 

The NLU unit is responsible for understanding users’ intent and extracting
related information that they enter so that the NLG unit can respond appropriately \citep{gupta2019casa}.
In this article, we are going to propose a model to not only detect the intention
of users but also check if their queries are in the domain of the chatbot’s
defined task and then cluster those unseen queries to map them to a pseudo label,
so we can retrain our model to cover a broader domain.

To clarify the problem, assume a customer who wants to book a train ticket from an assumptive origin to an assumptive destination. 
Then the customer may say something like "Book me a train ticket from my city to another city for 15th June at 2 pm." to the chatbot. 
If booking a train ticket had been defined for the chatbot as one of its tasks, everything
would go on as the customer wants until riching the goal (booking the
train ticket). However, what if the chatbot is just responsible for booking
airplane tickets? Some scenarios could happen. In the first scenario,
the chatbot would think that it should book a ticket for a flight
on 15th June at 2 pm; because it cannot distinguish defined (seen)
utterances from undefined (unseen) ones. In another scenario, the
chatbot can make users confirm the chatbot's predictions. It may
be a good idea, but repetition of that may be tedious and take time
for the user. Adding Out of Domain (OOD) intent detection
ability to the chatbot can prevent mentioned problems.

After solving OOD intent detection, intent discovery methods
can be used to detect the most frequent OOD utterances and put
them in clusters. The intent detection model can be retrained to
cover more intents by giving pseudo labels to these clusters. To
clarify the intent discovery problem, suppose that the defined
chatbot can detect OOD intents. So, it can detect the sentence above
as a novelty. We can store these unseen utterances using intent discovery and apply clustering methods. Then it would be
more feasible and less time-consuming to expand our chatbot to
cover more services \citep{congying2018zero}.

Although many good studies have been done in these fields, some challenges remain.
Some of these remaining challenges can be categorized as follows:
\begin{itemize}
  \item \textbf{Curse of Dimensionlity}: This challenge is related to OOD intent detection and intent discovery. Many methods use density or distance metrics (like euclidean distance) for clustering and OOD detection, which may not perform well on high-dimensional data \citep{xia2015effectiveness}.
  \item \textbf{Confusing Similarity}: This problem involves utterances with different intents but common words and similar grammatical structures. For example, two utterances, “I want a flight ticket” and “I want a cinema ticket,” have common words and similar sentence structures. These kinds of inputs can confuse the model to understand the intent correctly. This problem is more common in chatbots that give a specific service in a domain, for example, giving ground transport services but not giving services related to buses.
  \item \textbf{Distribution Dependency}: Many statistical methods have been introduced to detect OOD data \citep{gen2021}, but most have assumptions on how original data distribution should be. It is tough to talk about the input distribution or say it will always follow a specified distribution. So, using methods with assumptions on distribution may have a good performance on training data, but the performance would decrease in serving.
  \item \textbf{Parameter Selection}: Clustering results are highly dependent on the value of hyperparameters \citep{fan2020hyperparameter}. Setting a correct value for some parameters, like the number of clusters, is critical and needs previous knowledge about data.
\end{itemize}
This article will propose an architecture that detects and discovers new intents.
A \textbf{V}ariational \textbf{A}uto-\textbf{E}ncoder (VAE) is used to solve mentioned challenges for OOD intent detection.
VAEs can detect inputs with novel contexts independent of their distribution.
Also, they perform well on high-dimensional data.
We use a kernel-PCA algorithm for intent discovery to overcome the problems that high-dimensional data may cause.
After that, HDBSCAN will be used for clustering.
HDBSCAN does not ask for the number of clusters as a hyperparameter.
We also set other hyperparameters of HDBSCAN using information lying inside train data (without making any assumptions on utterances with OOD intents).
Results show that the proposed method has performed well in English and Persian.

The next section (\ref{relatedworks}) will briefly review previous works on OOD intent detection and intent discovery.
Section~\ref{methodology} contains details of the proposed method.
Section~\ref{experiments} is about datasets, implementation details and results.
Lastly, section~\ref{conclustion} will summarize what had been done and
will also talk about what actions can be taken in the future
to improve and fix the weaknesses of this work.

\section{Related Work}\label{relatedworks}
\noindent The focus areas of this research are OOD intent detection and intent discovery.
Hence, this section will be divided into two sections to cover both.

\subsection{OOD Intent Detection}\label{ood}

\noindent The solutions to detecting OOD inputs can be widely divided into different groups \citep{gen2021}. 
To solve the problem of OOD intent detection, 
some use the approach of generating pseudo-OOD data to make the model able to detect OOD inputs. 
The process of generating pseudo-OOD inputs can be done using adversarial learning or Autoencoders \citep{10.1145/3477314.3507089,marek2021oodgan,zheng2020out,ryu2018out}. 
Generating pseudo-OOD inputs using a combination of in-domain data representations is another approach \citep{zhan-etal-2021-scope}. 

Although pseudo-OOD samples may help classifiers find a sub-optimal boundary for in-domain data, 
they may fail to cover the whole boundary and also suffer from high complexity \citep{vernekar2019out}. 
Hence, some other work has been done to create models that detect OOD inputs without using OOD samples during training. 
The variety of these kinds of approaches is wide.
One approach can be using density-based anomaly detection methods like LOF \citep{yan2020unknown,lin-xu-2019-deep}. 
Another approach is to use distance-based methods. 
For example, the distance of the input can be calculated from the representative of each of the in-domain classes and considered as OOD input if they exceed a threshold \citep{podolskiy2021revisiting}.
Another approach to detecting OOD inputs is assuming a geometrical shape to find a boundary for each in-domain class \citep{zhang2021deep}. 
Input will be considered as OOD if it lies outside of defined boundaries.
Also, working on the representations is a way to overcome the problem of recognizing OOD inputs. 
An example of this approach is using contrastive learning \citep{jin2022towards}.
The last example of approaches to detect OOD samples without generating them during training is to work with the probability distribution of inputs\citep{10.1145/3397271.3401318}.

\subsection{Intent Discovery}

\noindent Intent discovery involves discovering intents lying behind inputs with unknown intents and is assumed as an unsupervised task. 
In some research, this problem is addressed by making assumptions about the structure of the intents. 
For example, the grammatical role of each input word is found using a named entity recognizer. 
Then the intents will be discovered by combining some words with specific roles \citep{10.1145/3366423.3380268,zeng2021automatic}.

Another approach to solving the intent discovery problem is applying straightforward clustering approaches to the data. 
These approaches usually need a good representation of inputs and depend on the clustering method's hyperparameter selection \citep{padmasundari2018intent,zhang2021discovering}.

As we saw, the previous works have also referred to challenges like distribution dependency \citep{zhan-etal-2021-scope}, the curse of dimensionality \citep{padmasundari2018intent}, 
semantical similarity between different intents \citep{lin-xu-2019-deep}, and parameter selection \citep{zhang2021discovering}. 
In the following sections, we will discuss our proposed method, which tries to solve these challenges.

\section{Methodology}\label{methodology}

\begin{figure*}[ht!]
	\centering
	\includegraphics[width=\textwidth]{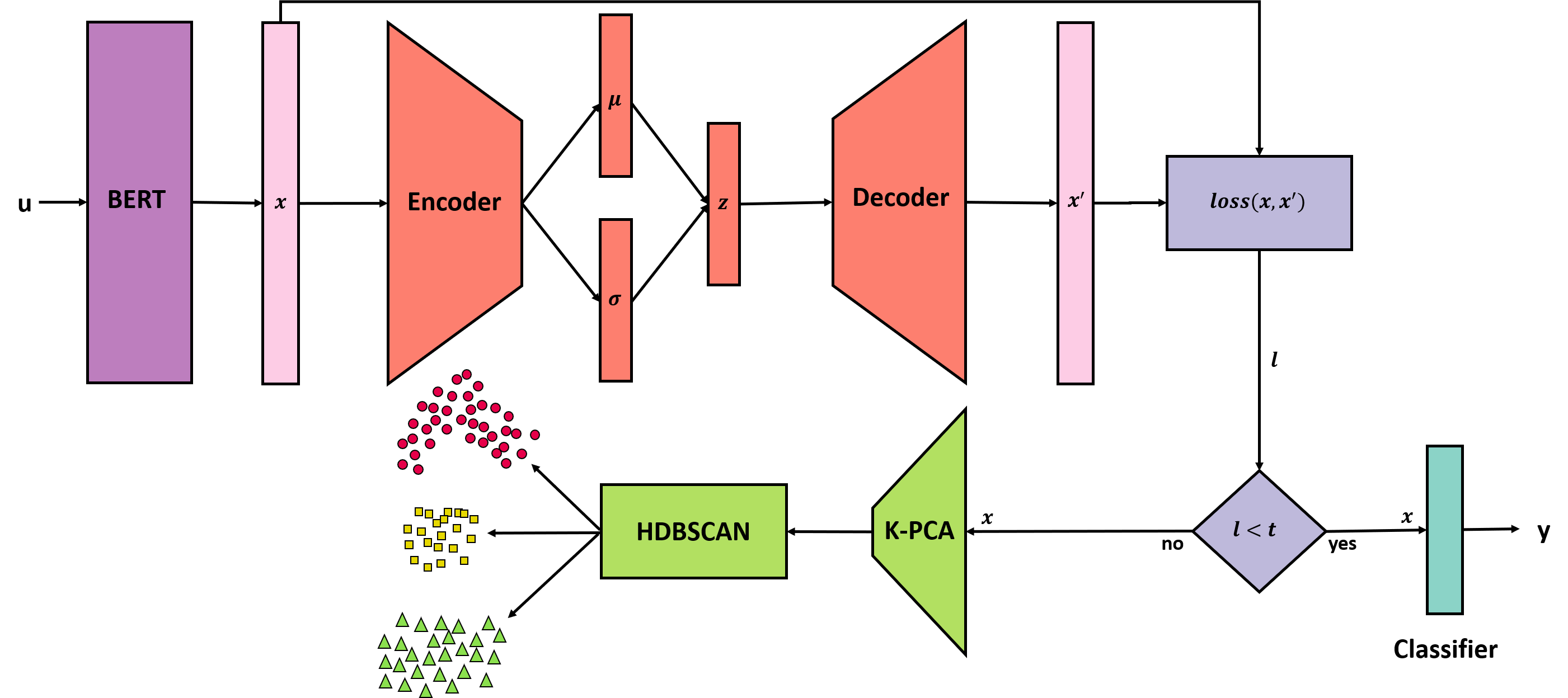}
	\caption{This figure shows the whole flow of the proposed architecture. 
			In the first step, the fine-tuned BERT encoder gets utterance $u$ and passes its representation $x$ to the VAE. 
			Then, the reconstruction $x'$ is generated, and the loss $l$ between $x$ and $x'$ is computed. 
			Suppose $l$ is lower than a threshold $t$. 
			In that case, $u$ will be considered in-domain input, 
			and the related representation $x$ is passed to the simple one-layer neural network classifier to predict its label. 
			Otherwise, $x$ will be passed to the K-PCA dimensionality reduction component, 
			and the HDBSCAN clustering algorithm will predict the resulting vector's pseudo label (or cluster).}
	\label{fig:Figure 1}
  \end{figure*}

\noindent Figure \ref{fig:Figure 1} shows the whole proposed architecture.
The following sections give a detailed description of the proposed method's steps.

\subsection{Problem Formulation}

\noindent Given a set \begin{math}U = \{u_{1}, u_{2}, ..., u_{n}\}\end{math} of utterances with known intents and
\begin{math}U^{\prime} = \{u_{1}^{\prime}, u_{2}^{\prime}, ..., u_{m}^{\prime}\}\end{math} of utterances with unknown intents,
the task can be formulated as follows:

\begin{gather*}
  input\ :\ u\ |\ u\ \in\ U\ \vee\ u\ \in\ U^{\prime} \\
  output\ :\ y\ s.t.\ \bigg\{_{pseudo\ label\,\quad if\ u\ \in\ U^{\prime}}^{y\ \in\ I\,\qquad if\ u\ \in\ U}
\end{gather*}

Where $I$ is the set of known intents and y is the output produced by our model. 
Based on this definition, the model will return the corresponding label if the input lies inside known intents—otherwise,
a pseudo label, which is the related cluster label will be returned.

\subsection{Sentence Representations}

To have an efficient representation of sentences, we use BERT encoder \citep{vaswani2017attention} for English and ParsBERT encoder \citep{farahani2021parsbert} for Persian dataset.
We first fine-tune BERT on our in-domain training data. A previous research done by Podolskiy et al. shows that
finetuning makes extracted representations more efficient \citep{podolskiy2021revisiting}.
Given an utterance of length $L$, BERT will return embeddings as $[CLS, e_{1}, ..., e_{l}] \in \mathbb{R}^{(L+1)*h}$,
were $h$ is the hidden layer size and $CLS$ is a representative for whole utterance which is mostly used for classification.
We will use $CLS$ for OOD intent detection and intent discovery, too.

\subsection{OOD Intent Detection}\label{id}

\subsubsection{Variational AutoEncoder}
VAEs are a specific kind of AutoEncoders.
An AE is made of 2 parts: an encoder and a decoder.
The encoder transforms input to a latent space with a lower dimension.
On the other side, the decoder tries to reconstruct the input data, given latent representation.
The decoder guarantees that the resulting representation in latent space is good enough that the decoder can reconstruct the original input data with a low error.

Instead of finding a lower-dimensional representation of the given input,
VAE tries to learn a distribution by estimating mean and standard deviation using neural networks.
Learning the distribution happens by adding a regularizer to AE's loss function.
The regularizer is a Kullback-Leiber divergence term.
This term makes the resulting distribution in latent space the same as a specific distribution.
We usually set this specific distribution to be Gaussian.

Equation \ref{eq:1} shows the loss function for training VAE.
\begin{equation}
  \begin{aligned}
    L = \bigg[ & - \sum_{i=1}^{n} (y_{i}\log\hat{y}_{\theta,i} + (1 - y_{i})\log(1 - \hat{y}_{\theta,i})) \\
    & + D_{KL}(q(z)_{\phi} || p(z)_{\theta})\bigg] \label{eq:1}
  \end{aligned}
\end{equation} 

First term of this formula is a crossentropy that calculates the loss between the input and reconstruction.
The second term is Kullback Leiber divergence between estimated ($q(z)_{\phi}$) and specified ($p(z)_{\theta}$) distributions.
This loss function should be minimized during training.

\subsubsection{OOD Intent Detection using VAE}

AEs (and so VAEs) can perform well in the training domain \citep{noroozi2016unsupervised,NIPS2015_aa169b49}.
It means they can find a good representation of in-domain data and reconstruct them with a low loss.
This property makes AEs and VAEs excellent candidates for detecting out-of-domain inputs.
We first train the VAE on our in-domain training data.
After training, VAE would be able to reconstruct in-domain inputs with a low loss.
So at inference time, if we get a high value for loss, the input will be considered an OOD input.
We set a threshold on loss values to identify OOD inputs
(if the computed loss is greater than the threshold, the input is OOD).

The reason for using VAE instead of AE is the regularizer in the VAE loss function.
The representations generated by VAE in latent space are smoother than representations generated by vanilla AE.
So the representations learned by VAE perform better for OOD Intent Detection \citep{an2015variational}.

\subsection{Intent Discovery}

\subsubsection{Dimensionality Reduction}

As was noted in earlier sections, the curse of dimensionality has been a challenge to clustering. 
In other words, as the dimension of representations increases, the significance of distance or similarity measures decreases \citep{10.1007/3-540-44503-X_27}.

Kernel-PCA is a variant of PCA, which has been highly used for dimensionality reduction \citep{10.1007/BFb0020217}. 
PCA seeks to find the linear combination of dimensions that explains the most variance in the data. 
The process starts by standardizing the data. The covariance matrix is computed in the second step to find the dimensions' correlation. 
Then, the Eigenvectors and Eigenvalues of the covariance matrix will be computed, and the Eigenvectors should be sorted decreasingly concerning their related Eigenvalue. 
The first p vectors should be chosen as principal components if we want to transform data to a p-dimensional space.

Kernel-PCA has a preliminary step to all those related to PCA: 
applying a kernel function to transform data to a higher dimensional space where the data will be linearly separable. 
Hence, kernel-PCA can capture nonlinear relations between dimensions, too.

\subsubsection{Clustering}

\noindent Setting the best values for the hyperparameters of a clustering algorithm is critical. 
However, finding those best values is challenging and sometimes needs background knowledge of data \citep{van2017constraint}.
In real-world scenarios, specifically OOD intent discovery, where we do not know the incoming data, 
it is hard to find the best set of values for hyperparameters. 
One of these parameters is the number of clusters. 
Although some methods like elbow can help us estimate the number of clusters, they need to be more accurate.

We found HDBSCAN\footnote{Hierarchical Density-Based Spatial Clustering of Applications with Noise} to be an efficient solution for these issues. 
HDBSCAN is an expansion of the DBSCAN method \citep{10.1007/978-3-642-37456-2_14}, 
which likewise employs density information to find clusters, 
but HDSBCAN is more able to deal with clusters with varying densities. 
HDBSCAN operates by generating a hierarchy of clusters, where a single point in the hierarchy represents each cluster. 
The method begins by identifying a set of core points with a sufficient number of adjacent points within a predetermined distance (called the eps parameter). 
The first level of the hierarchy is then created by grouping these core points. 
The algorithm then grows the clusters in the hierarchy iteratively by adding points that are not core points 
but are within a specific distance (called the min samples parameter) of a core point. 
A point close to multiple core points is added to the cluster with the highest density. 
This procedure is repeated until each data point is allocated to a cluster \citep{McInnes2017}. 
HDBSCAN is also very effective because the user is not required to define the number of clusters in advance. 

We have used the in-domain training data to set HDBSCAN hyperparameters (\emph{eps} and \emph{min\_samples}). 
Further information, the set of the values of hyperparameters is selected using in-domain training data, 
which results in the best performance for clustering of in-domain training data.

\section{Experiments}\label{experiments}

\subsection{Implementation Details}

\noindent We have used the $CLS$ tag of BERT-base and ParsBERT as the utterances' representations. 
Both encoders (Persian and English) provide us with 756-dimensional vectors. 
They have also been finetuned on the in-domain training data. 
The model with the highest score on validation data on each epoch has been selected to prevent overfitting. 
The VAE has an encoder with three layers, and the layers have 521, 256, 128, and 64 nodes in order. 
Also, the latent representation's size is 32 (the impact of different sizes for the latent representation is shown in the following sections). 
All these implementations have been done using Tensorflow in python.

For kernel-PCA implementation, the sklearn python library was used. 
The data was transformed to a 2-dimensional space using a polynomial kernel function. 
So, we used the euclidean distance measure was used within HDBSCAN\footnote{hdbscan.readthedocs.io}. 
The cluster selection method was also selected to be $leaf$ to prevent generating large heterogeneous clusters.

It is also important to note that none of the out-of-domain data was used during the training process and was only used for evaluating the proposed method.

\subsection{Datasets}\label{datasets}

\begin{table}[t!]
  \centering
  \caption{Randomly selected intents as out-of-domain}
  \label{tab1}
  \begin{tabular}{cc}
      \toprule
      Dataset  & Out-of-Domain Intents \\ \hline
      \midrule
      ATIS \&     & airline, meal, airfare, day\_name,        \\
      Persian-ATIS & distance \\ \hline
      SNIPS          & GetWeather, BookRestaurant        \\
      \bottomrule
  \end{tabular}
\end{table}

\noindent The proposed method is evaluated on three different datasets (one Persian and two English).
These datasets are as follows:
\begin{itemize}
  \item \textbf{ATIS} \cite{hemphill-etal-1990-atis}: ATIS is a dataset containing text utterances in air transport services with 26 different classes. As all utterances in ATIS are related to the same domain, different classes have some semantic similarities. This similarity makes ATIS more challenging for OOD intent detection and Intent Discovery.
  \item \textbf{SNIPS} \cite{coucke2018snips}: Unlike ATIS, SNIPS contains utterances with different domains. SNIPS dataset contains five different classes, and the similarity between the utterances of those classes is low.
  \item \textbf{Persian-ATIS} \cite{akbari2023persian}: The Persian version of ATIS has the same characteristics as the original ATIS. The only difference is the language\footnote{github.com/Makbari1997/Persian-Atis}.
\end{itemize}

To prepare these datasets for OOD Intent Detection and Intent Discovery tasks,
we randomly select some classes of each dataset as Out-of-Domain data.
Table~\ref{tab1} shows our grouping into in-domain and out-of-domain.

\subsection{Baselines}

\noindent To compare the results of our proposed method, we chose some research with similar approaches to ours.
Choosen research as baseline for OOD Intent Detection are as follows:
\begin{itemize}
  \item \textbf{LMCL + LOF} \citep{lin-xu-2019-deep}: This method uses the Large Margin Cosine Loss that makes the feature extraction layers - LSTMs here - 
                                                    extract features that maximize the interclass similarity and minimizes intraclass similarity. 
                                                    Then, those features would be passed to a Local-Outlier Factor to distinguish between in-domain and OOD inputs. 
                                                    In the original paper, they used glove \citep{pennington2014glove} word embeddings to get utterances' representations. 
                                                    As the glove Persian word embeddings do not have the same dimension as its English version and may show weak performance, 
                                                    this method is also implemented using fasttext \citep{DBLP:journals/corr/JoulinGBM16} pre-trained word embeddings, which perform better in Persian.
  \item \textbf{KLOOS} \citep{10.1145/3397271.3401318}: KLOOS uses Kullback-Leiber divergence to detect OOD inputs. 
                                                      This method uses a fully-connected neural network with a softmax activation function to find a distribution for each input word. 
                                                      The distribution shows the probability of a word belonging to each intent. 
                                                      Then, the KL divergence of consecutive words is calculated and passed to a binary classifier to decide whether the input is OOD. 
                                                      This approach also uses fasttext word embeddings. 
                                                      They have used different classifiers in the paper. 
                                                      Here we use KLOOS with logistic regression classifier, which has been mentioned as one of the classifiers with the best results in the paper. 
  \item \textbf{ADB} \citep{zhang2021deep}: This research presents a new method that assumes sentences with specific intent lie in a bounded spherical area in space. 
                                          They set the centers of spherical areas with the means of sentence representations of each class, 
                                          and then their radius is found using a soft-plus activation function. 
                                          The euclidean distances of each input to centers should be calculated to detect the related intent. 
                                          If the calculated distance exceeds all radiuses, the input will be considered OOD. Otherwise, 
                                          the related intent would be the intent with the nearest center. 
                                          ADB (Adaptive Decision Boundary) uses a fully connected neural network on top of a BERT-base encoder to get utterances' representations. 
                                          They apply a mean pulling on the BERT's outputs and then pass them to the fully connected neural network.
                                          ParsBERT has also been used here to apply ADB on the Persian dataset.
\end{itemize}
For Intent Discovery, we chose \textbf{Deep Aligned Clustering} \citep{zhang2021discovering}.
Deep Aligned Clustering uses a deep neural network to discover intents. 
In this approach, the deep neural network is trained based on the pseudo labels generated by a k-means clustering algorithm. 
They also proposed a method to find the best value for k. At first, k is initialized by a large number (for example, twice the number of clusters). 
Then, k-means is applied to the data, and at each iteration, some unconfident clusters are merged with the nearest clusters. 
The condition for a cluster to be determined as unconfident is to have a smaller number of members than a defined threshold.
This approach uses the same way of computing representations as \textbf{ADB}.

Also, it is crucial to note that there is no baseline for OOD Intent Detection and Intent Discovery in Persian.
Evaluating the proposed method and baselines in Persian and English at the same time is a way that can inform us of how well our method is performing.

\subsection{Metrics}

\noindent As the proposed approach consists of supervised and unsupervised methods, we have to evaluate it using different metrics.
We have used following metrics for OOD Intent Detection:
\begin{itemize}
  \item \textbf{F1-score} : ATIS (and so Persian-ATIS) is an imbalanced dataset. 
                          Furthermore, the number of in-domain and Out-of-Domain data are also often imbalanced. 
                          So we use the f1-score (macro and micro), which is a better metric for problems with imbalanced data.  
  \item \textbf{AUC-ROC} : The ability of the proposed method to detect OOD intents highly depends on the chosen threshold. 
                          AUC-ROC evaluates the performance of a model to distinguish true positives and negatives concerning different thresholds. 
                          Using AUC-ROC will inform us about the model's performance using different thresholds.
\end{itemize}
Intent Discovery evaluation metrics are as follows:
\begin{itemize}
  \item \textbf{NMI} : This metric calculates the amount of reduction of entropy inside clusters. In other words, higher NMI means lower entropy concerning labels inside a cluster.
  \item \textbf{ARI} : ARI compares the similarity of two clusterings. Here, we take ground-truth labels as one of the clusterings.
  \item \textbf{ACC} : Accuracy for clustering is the same as classification accuracy, with some differences. Accuracy for clustering should find a mapping between cluster labels and ground-truth labels.
\end{itemize}

\subsection{Results}

\begin{table*}[t!]
	\centering
	\caption{F1-score for OOD Intent Detection as binary classification}
	\label{tab:Table 2}
	\begin{tabular}{ccccccc}
	  \toprule
	  & \multicolumn{2}{c}{SNIPS} & \multicolumn{2}{c}{ATIS} & \multicolumn{2}{c}{Persian-ATIS} \\
	  & macro F1 & micro F1 & macro F1 & micro F1 & macro F1 & micro F1 \\ \hline
	  \midrule
	  glove  LMCL & 76.08 & 87.57 & 65.2 & 65.77 & 9.99 & 10 \\ \hline
	  fasstext + LMCL & 77.7 & 89.39 & 63.48 & 64.23 & 17.81 & 18.45 \\ \hline
	  fasstext + KLOOS & 47.17 & 89.3 & 36.6 & 57.72 & 35.03 & 55.7 \\ \hline
	  BERT + ADB & 62.79 & 73.35 & 83.73 & 83.74 & 42.15 & 47.97\\ \hline
	  \textbf{BERT + VAE} & \textbf{92.32} & \textbf{96.9} & \textbf{86.79} & \textbf{87.15} & \textbf{79.03} & \textbf{79.67} \\ \hline
	\end{tabular}
  \end{table*}

  \begin{table*}[t!]
	  \centering
	  \caption{F1-score for OOD Intent Detection as multi-class classification}
	  \label{tab:Table 3}
	  \begin{tabular}{ccccccc}
		\toprule
		& \multicolumn{2}{c}{SNIPS} & \multicolumn{2}{c}{ATIS} & \multicolumn{2}{c}{Persian-ATIS} \\
		& macro F1 & micro F1 & macro F1 & micro F1 & macro F1 & micro F1 \\ \hline
		\midrule
		glove  LMCL & 77.38 & 87.54 & 21.09 & 65.52 & 2.3 & 9.86 \\ \hline
		fasstext + LMCL & 74.08 & 89.38 & 18.94 & 61.54 & 3.28 & 16.61 \\ \hline
		fasstext + KLOOS & 15.73 & 89.3 & 5.72 & 57.72 & 6.23 & 54 \\ \hline
		BERT + ADB & 71.93 & 73.31 & 78.83 & 83.5 & 25.51 & 40.73\\ \hline
		\textbf{BERT + VAE} & \textbf{89.58} & \textbf{96.85} & \textbf{79.38} & \textbf{86.83} & \textbf{79.03} & \textbf{79.68} \\ \hline
	  \end{tabular}
  \end{table*}
  
  \begin{figure*}[ht!]
	\centering
	\includegraphics[scale=0.4]{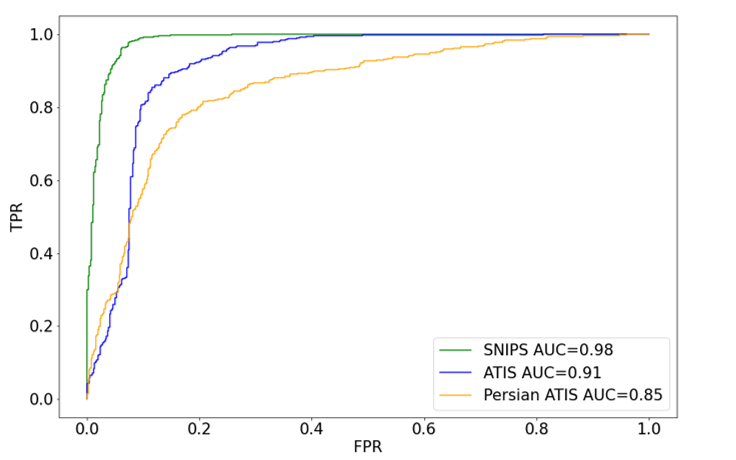}
	\caption{ROC curve and AUC value for all dataset; Results show that proposed method can separate true negatives and positives
			well on different thresholds.}
	\label{fig:Figure 2}
  \end{figure*}
  \begin{figure*}[ht!]
	\centering
	\includegraphics[width=\textwidth]{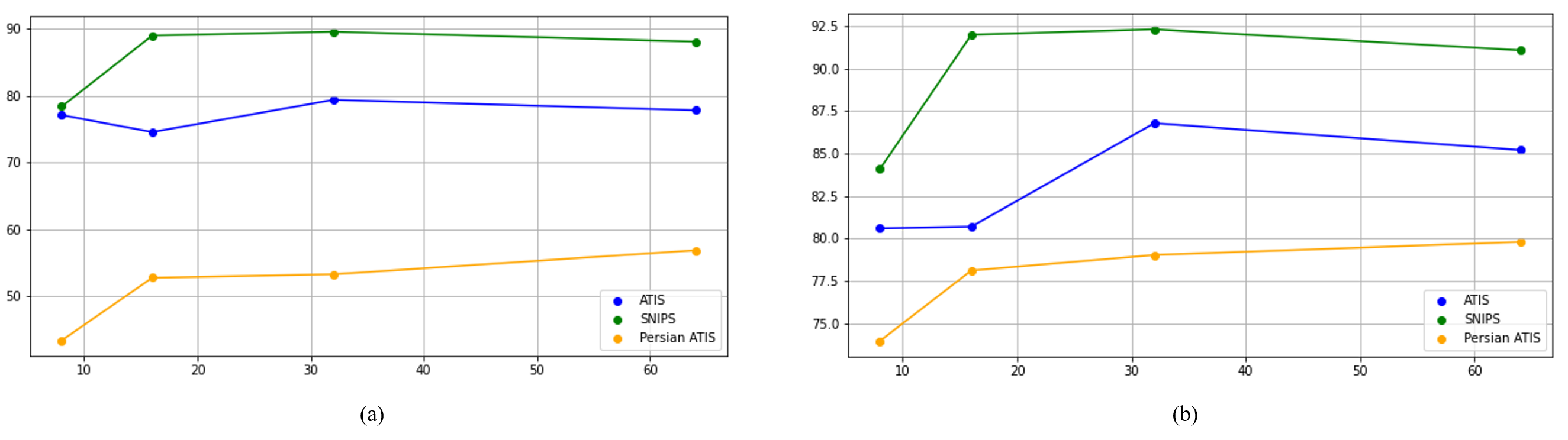}
	\caption{The effect of changes in the dimension of the representations in hidden space. x-axis is dimension and y-axis is the
			macro F1-score. Also, (a) is showing the effect in multi-class classification and (b) is showing the effect in binary classification.}
	\label{fig:Figure 3}
  \end{figure*}

\subsubsection{OOD Intent Detection}

Tables~\ref{tab:Table 2} and~\ref{tab:Table 3} show the results of applying baselines and the proposed method on three datasets. 
Table \ref{tab:Table 2} shows F1-score when the goal is to detect whether the input is OOD (same as a binary classification). 
Table \ref{tab:Table 3} shows F1-score when the model also predicts the intent of the in-domain inputs. 
The results show that our proposed method can outperform baselines in both scenarios.
The high value for the macro F1-score shows that our method can distinguish different classes well, even for imbalanced datasets (ATIS \& Persian-ATIS). 
As was noted in \ref{datasets}, ATIS (and so Persian ATIS) is a more challenging dataset since OOD intents overlap with in-domain intents semantically.
SNIPS has unrelated intents, and the semantical overlapping is low, so the model's performance is much better.

Since the VAE's ability to detect anomalies depends on the threshold, 
It is essential to analyze its performance concerning different values as the threshold. 
This analysis is done by computing the AUC value and ROC curve, which can be seen in Figure \ref{fig:Figure 2}. 
The figure and the value of the AUC of each dataset show that our method can separate true positives and true negatives using different thresholds.

One of the other factors that can affect the performance of the VAE is the size of the dimensions in the latent space. 
As shown in Figure \ref{fig:Figure 3}, 32 has the best performance among values 8, 16, 32, and 64.
The decrease in performance by reducing the size of the hidden state is that the resulting representation in latent space loses much valuable information. 
On the other side, increasing the dimension of latent space lowers the effect of gradient on the first layers, so they will not be able to extract good features.

\begin{table*}
	\centering
	\caption{Results of evaluating the proposed method for Intent Discovery (clustering)}
	\label{tab:Table 4}
	\begin{tabular}{cccccccccc}
	  \toprule
	  & \multicolumn{3}{c}{SNIPS} & \multicolumn{3}{c}{ATIS} & \multicolumn{3}{c}{Persian-ATIS} \\
	  & ACC & NMI & ARI & ACC & NMI & ARI & ACC & NMI & ARI \\ \hline
	  \midrule
	  Deep Aligned Clustering & 39.74 & 28.23 & 13.1 & 19.02 & 25.31 & 5.58 & 15.28 & \textbf{17.53} & 4 \\ \hline
	  \textbf{k-PCA + HDBSCAN} & \textbf{74.95} & \textbf{45.47} & \textbf{59.23} & \textbf{89.01} & \textbf{63.68} & \textbf{74.94} & \textbf{65.49} & 13.98 & \textbf{11.97}\\ \hline
	\end{tabular}
  \end{table*}
  \begin{table*}
	\centering
	\caption{The ability of methods to find different available intents in OOD utterances}
	\label{tab:Table 5}
	\begin{tabular}{ccccccc}
	  \toprule
	  & \multicolumn{2}{c}{SNIPS} & \multicolumn{2}{c}{ATIS} & \multicolumn{2}{c}{Persian-ATIS} \\
	  & $k$ & $k^*$ & $k$ & $k^*$ & $k$ & $k^*$ \\ \hline
	  \midrule
	  Deep Aligned Clustering & 7 & 2 & 23 & 5 & 20 & 5 \\ \hline
	  k-PCA + HDBSCAN & 3 & 2 & 3 & 5 & 3 & 5 \\ \hline
	\end{tabular}
  \end{table*}
  \begin{figure*}[ht!]
	\centering
	\includegraphics[scale=0.4]{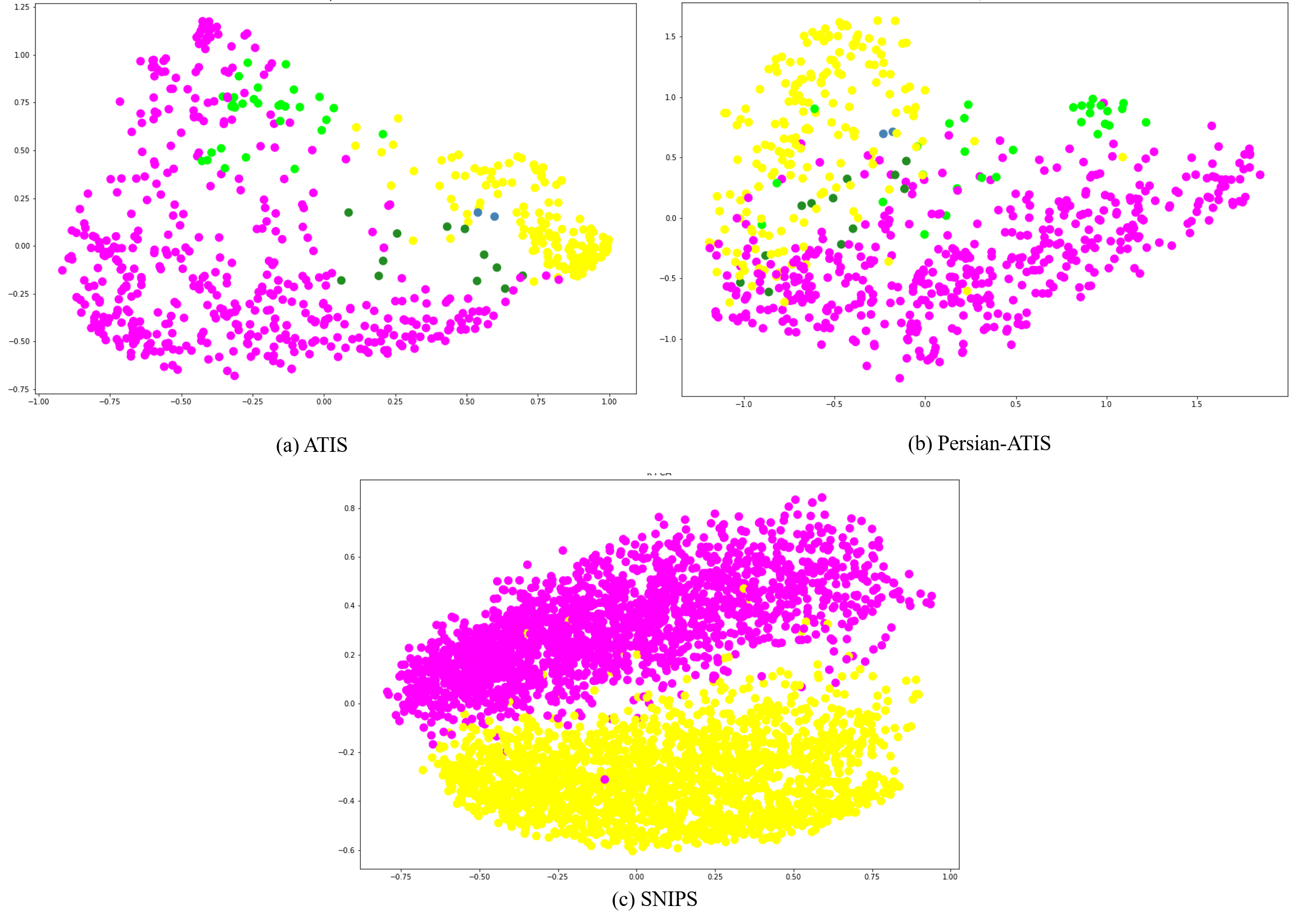}
	\caption{Quality of representations of OOD utterances after applying kernel-PCA}
	\label{fig:Figure 4}
  \end{figure*}

\subsubsection{Intent Discovery}

Table \ref{tab:Table 4} shows the results of the intent discovery part of the proposed method. 
Results show that using kernel-PCA and HDBSCAN together has better performance for clustering. 
The performance of the model is much lower in Persian. The reason may be that the pre-trained version of BERT-base in Persian needs to be stronger than the English version.

To understand how our method is performing to discover intents lying in OOD utterances, we can look at Tabe \ref{tab:Table 5}.
Based on the results, our proposed method and the way of selecting parameters are more successful in discovering different intents lying in OOD utterances.
Although the number of discovered intents ($k$) is not equal to the ground-truth number of intents ($k^*$), it is closer than the baseline predicting $k$.

The visualization of sentence representations after applying kernel-PCA will help us to understand the results of Table \ref{tab:Table 5} better.
Figure \ref{fig:Figure 4} visualizes the quality of representations of OOD utterances in 2-dimensional space after applying kernel-PCA.
As we can see, the quality is better on the SNIPS dataset. The reason is that the classes of SNIPS are not similar to each other as ATIS.

\section{Conclusion}\label{conclustion}

In this research, we have proposed a new hybrid method for OOD Intent Detection and Intent Discovery.
Also, this method is the first baseline for this problem in Persian.
Comparisons with baselines show that our approach can perform much better even on ATIS, where its utterances of different classes have similar and near semantics.
The need to set the threshold manually is a con of the proposed method.
In future works, we will try to make this approach end-to-end.
We can also try to generate some pseudo-OOD utterances using VAE.
Also, Variational Deep Embedding can be used for Intent Discovery.

\bibliographystyle{unsrtnat}
\bibliography{references}  






\end{document}